# Towards a Privacy-preserving Deep Learning-based Network Intrusion Detection in Data Distribution Services

*by Stanislav Abaimov*
*City, University of London*

## 1.1 Abbreviations

| | |
|---|---|
| API | Application Programming Interface |
| ATM | Air-traffic Monitor |
| CNN | Convolutional Neural Networks |
| DBNN | Deep Belief Neural Networks |
| DDS | Data Distribution Services |
| ICS | Industrial Control Systems |
| IDS | Intrusion Detection System |
| IoT | Internet of Things |
| IIoT | Industrial Internet of Things |
| LASSO | Least Absolute Shrinkage and Selection Operator |
| LSTM | Long Short-Term Memory |
| ML | Machine Learning |
| OMG | Object Management Group |
| PCA | Principal Component Analysis |
| PLC | Programmable Logic Controller |
| RFE | Recursive Feature Elimination |
| RNN | Recurrent Neural Networks |
| ROS | Robot Operating System |
| RTPS | Real-Time Publish Subscribe |
| RSMT | Robust Software Modelling Tool |
| SSH | Secure Shell |
| TLS | Transport Layer Security |
| VM | Virtual Machine |

## Abstract

Data Distribution Service (DDS) is an innovative approach towards communication in ICS/IoT infrastructure and robotics. Being based on the cross-platform and cross-language API to be applicable in any computerised device, it offers the benefits of modern programming languages and the opportunities to develop more complex and advanced systems. However, the DDS complexity equally increases its vulnerability, while the existing security measures are limited to plug-ins and static rules, with the rest of the security provided by third-party applications and operating system. Specifically, traditional intrusion detection systems (IDS) do not detect any anomalies in the *publish/subscribe* method. With the exponentially growing global communication exchange, securing DDS is of the utmost importance to futureproofing industrial, public, and even personal devices and systems.

This report presents an experimental work on the simulation of several specific attacks against DDS, and the application of Deep Learning for their detection. The findings show that even though Deep Learning allows to detect all simulated attacks using only metadata analysis, their detection level varies, with some of the advanced attacks being harder to detect. The limitations imposed by the attempts to preserve privacy significantly decrease the detection rate.

The report also reviews the drawbacks and limitations of the Deep Learning approach and proposes a set of selected solutions and configurations, that can further improve the DDS security.

## 2    Introduction

Data Distribution Services (DDS) are getting more widespread in Industrial Control Systems (ICS), IoT devices, and robotics. They are an evolutionary step for the ICT/IoT infrastructure and interconnectivity of the non-PC devices, both in flexibility and scalability. Flexibility of the protocol and cross-platform cross-language Application Programming Interface (API) allows developers to design and improve their projects, while conforming to a standard of the peer-to-peer middleware, combined with an out-of-the-box monitoring software. Should this emerging trend become more commonly used, securing it is of paramount importance to futureproofing industrial, public, and even personal devices and systems.

DDS software has a set of dependencies and requirements, such as programming libraries, an operating system, and a hardware more powerful than a traditional PLC used in the ICS networks. This complexity increases the vulnerability of such systems to advanced cyber attacks, especially when implemented by highly skilled malicious actors equipped with the zero-day vulnerabilities and exploitation kits.

As a solution to this challenge, Machine Learning (ML) has already proven to be the next step after the signature-based intrusion detection [1].  Furthermore, it has evolved in the recent decades, and is now confidently used for the development of the next generation Network and Computer Intrusion Detection Systems (IDSs). Deep Learning has shown itself as most suitable due to its flexibility and suitability for intrusion detection purposes [2].

Numerous academic articles have already been published on the applications of supervised [3][4], unsupervised [5][6], and reinforcement ML [7] for the network intrusion detection,  with cyber security being one of the relatively saturated areas of the ML application.

### 2.1    Problem statement

As DDS is a relatively new technology, the attacks over the DDS protocols are not detected by conventional intrusion detection systems. New approaches are needed, among which the most flexible variation of machine learning has to be used, which is deep learning [2].

According to the preliminary experiments with deep learning (see section 4.2.4), the available IDS systems do not necessarily detect malicious activity via the Real-Time Publish Subscribe (RTPS) protocols or any other DDS communication implementations. The DDS security is limited to only rule-based security leaving the system vulnerable to non-standard types of attacks.

The issues of privacy and confidentiality raise concerns from personal to the national and international levels. With the existing DDS vulnerabilities, ensuring their security may jeopardise privacy, as an IDS would require the access to network packet contents. The intrusion detection should be conducted with an anonymised traffic flow, creating constraints in detection.

The revealed scarcity of the publications in the area of the DDS security testifies for the need in a more extensive and deep research considering the growing adoption of DDS in the industries and IoT.

## 2.2 Goals and Objectives

The **goal** of the experiment is to contribute to the improvement of the security of the DDS, while preserving communication privacy, and explore the intrusion detection capacities of machine learning over the DDS protocols in the industrial setting.

The **objectives** include simulating malicious and benign activities in a virtual environment, and developing a proof-of-concept IDS for the *publish-subscribe* communication, using machine learning for the privacy-preserving data analysis. Upon completion of the experimental work, conclusions should be drawn on the viable use of supervised machine learning for intrusion detection over the DDS protocols.

To test the approaches to this problem solution, a series of experiments were conducted on the premises of the City, University of London. The report aims at describing the conducted experiments with the simulated attacks, their recording, preprocessing, and detection.

## 2.3 Scope and limitations

Machine learning has already been extensively applied to a variety of network intrusion detection protocols and attack vectors [2]. The scope of this report is limited to the intersection of the following three research areas that sill remain less represented and poorly covered in literature:

simulation of industrial communication in virtual environment using the DDS software and custom *publishers* and *subscribers*;

network intrusion detection through the traffic flow analysis;

supervised deep learning as a tool for data analysis, where the network traffic flow serves as input data.

## 2.4 Research methodology

Out of more than 150 papers on machine learning for the network intrusion detection, selected for this report from the open access journals, conference papers, and protocol security standards, only 23 were selected as matching the set research criteria. They were further revised to select only those related to the DDS security, and the machine-learning-based attack detection and methods directly related to the attacks simulated for the experimental work.

In addition, the findings of the previous research of the author in deep learning have also been taken into consideration for defining the experiment parameters.

The approach selected for the experiment is based on the network traffic flow analysis, which allows detection of behavioural anomalies in the network activity. Focused on the collection of metadata and avoiding the analysis of each packet, the method is more lightweight, and allows fast processing of larger volumes of data.

In addition, the traffic flow analysis allows to preserve privacy as it does not access the payload in the process.

This experiment used a virtual network of three Virtual machines (VM) (VirtualBox 6) with Ubuntu 18 and 2 virtual routers. Every VM had an identical setup (via cloning a VM), including the deployed RTI Connext DDS v6.0.1 and dependencies.

To date of the publication of this report, and as per the analysis of the publicly available sources, there is no known and available for the ML dataset that had recorded attacks via DDS within the conditions set in the experiment. The required dataset has been created for the experiment based on the simulated attacks - DoS, Clone, and "Malicious Subscriber", using the repurposed examples provided by RTI[1]. The final dataset contains up to 2000 malicious sessions of each attack, generated by automatic relaunch of *publishers* and *subscribers*. The traffic recordings (*.pcap) were analysed using the opensource IDS (Snort and Zeek) and VirusTotal that supports the traffic record analysis.

---

[1] Examples, RTI Community, RTI, last accessed 2020-06-27, https://community.rti.com/examples

Convolutional Neural Network (CNN) was selected for the experiment, as it performs better when analysing a multidimensional array of features using convolution; single CNNs and ensembles are used for the detection. Conversion of traffic records into traffic flow data was conducted using CICFlowMeter. The automated methods of feature selection were selected as offering broader functions than the manual ones.

The conducted experiment is based on the usecase, studied in the project titled *Aggregated Quality Assurance for Systems* (AQUAS).

The RTI Connext DDS was selected for the experimental work as it allows compatibility across multiple operating systems and programming languages. The test platform used a GNU/Linux operating system on each virtual machine.

In the three experimental scenarios, developed based on the previous research, the assumption is that the attacker has access to the executable of the *publisher* or *subscriber* and the *quality of service*, whether those executables are on a host device or in a virtual container. As modern operating systems allow the access to the list of running processes, their locations and arguments both on a hard drive and in RAM, the assumption is that the attacker can launch new instances of publishers and subscribers.

The practical implementation of each step will be presented in detail further in the report (see section 3).

## 2.5 Structure

This report consists of several chapters.

The Introduction has outlined the research area, setting goals and objectives.

Chapter 2 *Background* provides definitions and reviews the selected academic literature and the latest findings.

Chapter 3 *Experiment* presents the experiment and details approaches to running simulation exercises.

Chapter 4 *Discussion* targets drawbacks and limitations.

Conclusion presents findings, implications and defines further research areas.

## 3 Background

This Chapter provides definitions of the researched subjects, specifically of the Data Distribution Service, Deep Learning and Feature Selection, and reviews selected academic literature findings most relevant to the experiment objectives.

### 3.1 Data Distribution Service and its security

The Data Distribution Service (DDS) is an Object Management Group (OMG) middleware standard for a publish-subscribe pattern. A DDS Security Standard, developed by OMG[2], was updated to version v1.1 in 2018, and is the most recent version of this standard to date of the experiment. A variety of DDS implementations have been developed since the introduction of the standard: RTI Connext DDS, OpenSplice, OpenDDS, eProssima, etc.

---

[2] Object Management Group, last accessed 2020-06-29, https://www.omg.org/spec/DDS-SECURITY/About-DDS-SECURITY/

For a more convenient deployment in robotic systems, a Robot Operating System (ROS) was developed in 2007, with subsequent introduction of ROS2, and applied in IoT devices to simplify connectivity of applications. ROS includes several out of the box DDS implementations.

DDS can be implemented on multiple programming languages, e.g., Ada, C, C++, C#, Java, Python, Scala, Lua, Pharo and Ruby; and through different operating systems (Microsoft Windows and Linux) with the support of multicast and unicast network communications [8].

Almadani et al [9], studying the systems reliability, proposed an ambulance system model (E-Ambulance framework) using the RTI Connext DDS for the medical data transfer.

For the purpose of this research, the two types of programs used in the DDS got the following working definitions: the *publisher* is defined as a program that produces information (topics) and broadcasts to one or more subscribers; *subscriber* is defined as a receiving program, that can collect data only from those publishers it is subscribed to.

Modern DDS safety and security are based on the IoT security experience. However, DDS is still a relatively new concept (2001) as compared to the fully implemented and widely used protocols it is built upon - IP (1981), IoT appliance Telnet (1969) replaced by SSH (1995), and popular ICS protocols Modbus (1979) and S7comm (1995). The DDS security still remains less understood and yet to be explored.

The literature analysis showed a noticeable scarcity in the publications on the DDS security. White et al [10] investigated issues with the Real-Time Publish Subscribe (RTPS) protocol revealing a potential Denial of Service in it. Hamed et al [11] presented the DDS security top-level overview with specific attention to maritime cybersecurity. They distinguish three threat vectors: unauthorised subscription, unauthorised publishing, and tempering and replay. The paper concludes that DDS provides adequate security, however, a thorough analysis of the DDS capabilities has to be conducted prior to deployment.

Over the two decades of existence, the DDS implementations have accumulated a substantial amount of specific vulnerabilities. The *publisher* and *subscriber* are vulnerable to replay, injection, and information leakage. For example, with the direct access to a copy of the executable files, *publisher* and *subscriber* can be reverse-engineered. If the *publisher* executable is not encrypted and packaged, the names of Topics (and other variables) can be extracted with a "*strings*" bash command and manually analysed. Applications written on a scripting language (e.g., Python), can be accessed without a debugger for the information to be extracted and used further for the attack.

Additionally, some APIs have built-in functions, allowing to specifically request the data that can be used for further exploitation. For example, RTI Connext API has a *get_publishers* function[3], that can be used to enumerate *publishers* in the domain.

Furthermore, the DDS implementations can be exploited through misconfiguration of the deployed software. Michaud et al [12] provided an empirical study of five (out of 60) DDS security issues using RTI Connext DDS (v5). They review, similar to Hamed et al [11], the following possible attacks:
- Unauthorised subscription
- Unauthorised publishing
- Tampering and replay

Specifically, the vulnerable points of the RTI Connext DDS were identified by [12] as:
- Anonymous Subscribe and Republish Functionality
- OWNERSHIP_STRENGTH QoS Policy and EXCLUSIVE Data Ownership
- OWNERSHIP_KIND QoS Policy and SHARED Data Ownership
- LIFESPAN QoS Policy Causing Immediate Data Expiration

---

[3] Research, RTI, last accessed 2020-06-28, https://research.rti.com/examples/get-publishers

- LocatorList Environment Variable Causing Domain Misdirection During Participant Discovery

Friesen et al [13] reviewed the security of the Real-time Publish Subscribe and Data-Centric Publish Subscribe protocols, as well as the TLS and DTLS protocols applied to the DDS architectures. The DDS Security Plugins include:
- Authentication Service Plugin
- Access Control Service Plugin
- Cryptographic Service Plugin
- Logging Service Plugin

### 3.2 Deep Learning and Feature selection

Deep learning is popularly applied for intrusion detection and has a wide variety of tools and approaches [2]. Deep Learning is applied to such types of artificial neural networks as deep neural networks (DNN), deep belief neural networks, recurrent neural networks (RNN) and convolutional neural networks (CNN) [14]. All of them have been widely researched both by practitioners and scholars.

Gu et al [15] and then Khan et al [16] present comprehensive surveys on the advances in the CNN architecture and techniques. CNNs are known to be effective for the Network Intrusion Detections when analysing data with high dimensionality. However, any kind of infiltration attacks or anomaly detection would require time distributed analysis, and will have to use RNN, or even LSTM.

Feature selection is an essential process for the successful intrusion detection using any ML-based method, as carefully selected features guarantee higher learning rate and improve the performance of the learners. Datasets may have more features than needed for a specific task, with some of them being even harmful for the results. Their filtering, refinement, and selecting only those relevant for the task improves the accuracy and results in higher performance of the ML-based system.

There have been developed a set of methods [17], both manual and automated, to determine more impactful features. Manually they can be selected via trial and error, while automatic selection implies the use of Principal Component Analysis (PCA), Recursive Feature Elimination, LassoCV, etc.

Automated methods offering a wider variety of possibilities and functions have been selected as oppose to manual methods. Their results were compared (see Table 1 in section 4.3.3) and the most impactful features were further used for the experiments.

The following methods, being available in SciKit Learn[4] for the automated feature selection, have been used:
- Recursive Feature Elimination
- Least absolute shrinkage and selection operator (LassoCV)
- Univariate selection
- Feature Importance

*Recursive Feature Elimination* [18][5] (RFE) is a feature selection method that fits a model and removes the weakest feature (or features) until the specified number of features is reached. The importance calculations can be model based (e.g., the random forest importance criterion) or done through a more general approach that is independent of the full model [19].

---

[4] Feature Selection, SciKit Learn, last accessed 2020-06-29, https://scikit-learn.org/stable/modules/feature_selection.html

[5] Recursive Feature Elimination, SciKit Learn, Yellowbrick, last accessed 2020-06-29, https://www.scikit-yb.org/en/latest/api/model_selection/rfecv.html

*Least absolute shrinkage and selection operator* (LASSO) [20] "*is a regression analysis method that performs both variable selection and regularization in order to enhance the prediction accuracy and interpretability of the resulting statistical model*". It is acknowledged, that the LASSO method penalises all coefficients equally, even when the coefficients are large. However, unlike a superior method *Adaptive Lasso*, LASSO is a method that is available out of the box in many programming libraries, including SciKit.

*Univariate selection* is a selection of a single variable in the data set. "*Statistical tests can be used to select those features that have the strongest relationship with the output variable*"[6].

*Feature importance* property of the model is a method that yields the feature importance of each feature of the dataset. It calculates a score for each feature of the selected data, the higher the score the more relevant is the feature towards the desired output variable.[7]

The following chapters present the experimental work and its results.

## 4 Experiment

The approach selected for the experiment is based on the network traffic flow analysis in order to detect anomalies in the network traffic. Focused on the collection of metadata and avoiding the analysis of each packet, the method is more lightweight than the traffic analysis, and allows simultaneous processing of larger volumes of data.

In addition, the traffic-flow analysis allows to preserve privacy as it does not access the payload in the process.

### 4.1 Setup

This experiment used a virtual network of three Virtual machines (VM) (VirtualBox 6) with Ubuntu 18 and 2 virtual routers. Every VM had an identical setup (via cloning a VM), including the deployed RTI Connext DDS v6.0.1 and dependencies. Graphical interface was enabled to allow the use of Wireshark[8] for traffic recording. With the system-wide vulnerabilities and attack vectors been extensively researched, this experiment scope of the analysis is narrowed down to the RTI Connext DDS.

This experiment setup allowed the findings to be transferrable to the following usecases:
- UAVs (publishers) transfer data (GPS and telemetry) to ground control (Subscriber)
- Heartrate monitor (published heartrate every second)
- CPU monitor (publishes load and temperature every second)

During the initial setup, there were identified two aspects of functionality that can be used or misused. The first aspect is related to the multiple instances of the same publisher being launched; while the second one is linked to the execution parameters and arguments. An instance of a *publisher* or a *subscriber* can be launched with the injected execution arguments, that may result in data disclosure or even an arbitrary command execution.

---

[6] SciKit Learn Documentation, last accessed 2020-06-29, https://scikit-learn.org/stable/modules/generated/sklearn.feature_selection.SelectKBest.html#sklearn.feature_selection.SelectKBest

[7] SciKit Learn Documentation, last accessed 2020-06-29, https://scikit-learn.org/stable/auto_examples/ensemble/plot_forest_importances.html

[8] Documentation, Wireshark, last accessed 2020-06-28, https://www.wireshark.org/docs/

### 4.1.1 Multiple publishers

*Publisher*, as any program or script, is by default vulnerable to false data, code injection, and intentional or unintentional denial of service. Publisher executables are modified examples, provided in the RTI Connext DDS package.

Launching multiple instances of the same *publisher*, if left unchecked, can cause the same data to be repeatedly sent to subscribers. Launching repeatedly (with a script) a large number of *publishers* may overrun the maximum network bandwidth or, more likely, the local system memory.

Changing the input data is another way to inject code or cause data overflow. This data can be injected from the execution arguments, files, networks, hardware.

The experimental attempts to launch additional C++ instances were successful. This confirms, that under certain conditions it is possible to launch additional instances of *publishers,* with different topics from the original *publisher*, and still deliver messages to the same *subscriber*.

### 4.1.2 Execution arguments

Unintended arguments can be injected through the repeated execution of the *publisher* instances. A simple malicious script can inject those arguments from a fuzzing dictionary and automatically test the vulnerability of both the *publisher* and the *subscriber*.

Changing execution arguments happens at the system level and goes beyond the implementations of the software and can be used as an attack method against any programming language or operating system, as long as the execution command matches the operating system syntax.

## 4.2 Simulations

To the extent of our knowledge, there are no datasets available in the public domain, that meet the set requirements at the time of the experiment implementation. To resolve this issue and to **create** the required datasets, three different attacks were simulated - DoS, Clone, and Malicious Subscriber, using the repurposed examples, provided by RTI[9].

The traffic recorded during these simulated attacks, contained up to 2000 malicious sessions of each attack, generated by automatic relaunch of *publishers* and *subscribers*, that exchanged 50-500 numeric values twice a second. Both malicious and non-malicious *publishers* and *subscribers* are indistinguishable when the conventional methods are applied.

*Publishers* constantly broadcast data, which in this experiment are the GPS Coordinates – 51N 5279719, 0W 1024624[10]. They change and range up to the maximum delta of 0.015 (~1 km) from the nominal, emulating a square flight pattern of an autonomous UAV, patrolling the designated perimeter.

The data broadcasted by *publishers* can be of any type, e.g., the telemetry data between the IoT devices and a control server, that is a comparatively large data frame, or a set of frames, containing GPS and relative coordinates, home point coordinates, controller information, battery information, identification data, etc.

After the simulation, the traffic recordings (*.pcap) were analysed using opensource IDS (Snort and Zeek) and VirusTotal, that supports traffic record analysis.

Below is a more detailed presentation of the layout and data transfer directions in the virtual environment for the simulated attacks.

---

[9] Examples, RTI Community, RTI, last accessed 2020-06-28, https://community.rti.com/examples
[10] GPS coordinates of City, University of London

### 4.2.1 Denial of Service

During the "Denial of Service" (DoS) attack simulation (See Figure 1) a malicious *publisher* (10.0.5.6) is identical to the genuine *publisher* (10.0.5.5). Malicious *publisher* broadcasts topics to the *subscriber* (10.0.5.4) more frequently and with a significantly larger payload, i.e. maximum allowed number of characters per topic.

"DoS" setup can be as follows:
- VM1 (10.0.5.4) – *subscriber*, Wireshark
- VM2 (10.0.5.5) – *publisher* x5 – interval: 0.5sec – payload: GPS, telemetry
- VM3 (10.0.5.6) – *publisher* (DoS) – interval: none – payload: "A"*256 (maximum allowed length in the configuration file). It publishes topics with maximum size payloads with minimal available delay.

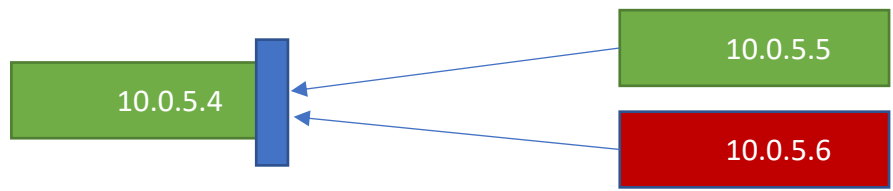

*Figure 1 Data transfer between virtual machines during "Denial of Service" flood attack*

### 4.2.2 Clone

In the "Clone" attack, the Clone (10.0.5.6, Figure 2) is a *publisher* that is identical to the genuine (10.0.5.5), but it is programmed to continuously publish topics with the same rate and with the false data.

"Clone" setup:
- VM1 (10.0.5.4) – *subscriber*, Wireshark
- VM2 (10.0.5.5) – *publisher* x5 – interval: 0.5sec – payload: "79", "80" or "81"
- VM3 (10.0.5.6) – *publishe*r (fake) – interval: 0.5sec – payload: "200" or "A"*256

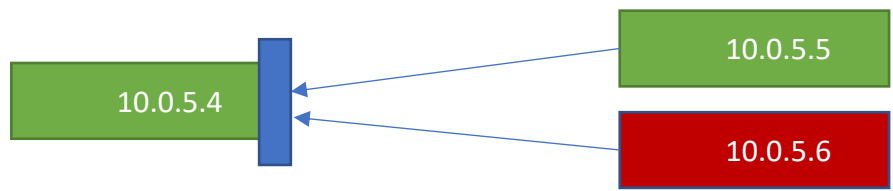

*Figure 2 Data transfer between virtual machines during "Clone" attack*

### 4.2.3 Malicious Subscriber

In the "Malicious subscriber" attack, the malicious *subscriber* (10.0.5.6, Figure 3) is a *subscriber* with the legitimate credentials and is identical to the genuine *subscriber* (10.0.5.5), launched by malware to collect data from the DDS. The *publisher* (10.0.5.4) broadcasts topics to any *subscriber* in the network. Mode of operation is selected in *QoS.xlm* to be "multicast". Initially, only the legitimate *subscriber* is active, and after some time a malicious *subscriber* is launched for a limited period of time, to collect the broadcasted data.

"Malicious subscriber" setup:
- VM1 (10.0.5.4) – publisher, Wireshark
- VM2 (10.0.5.5) – subscriber, 200 topics
- VM3 (10.0.5.6) – subscriber, 50-60 topics

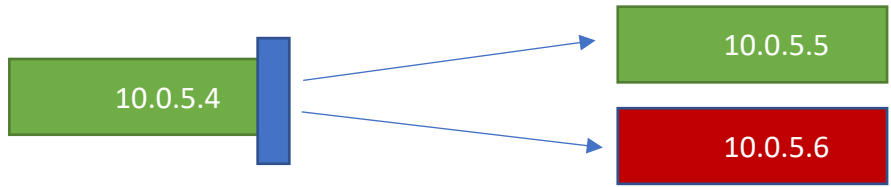

*Figure 3 Data transfer between virtual machines during "Malicious Subscriber" attack*

The presented attacks were recorded and saved in *.pcap format for further analysis.

### 4.2.4 Conventional attack detection

The opensource conventional IDS, Snort[11] and Zeek[12], were used to test for their capacity to recognise any of the above-mentioned attacks. Both IDS demonstrated that none of the attacks were detected. In addition, the files were uploaded to VirusTotal to confirm that the DoS flood was not detected either (see Figure 4).

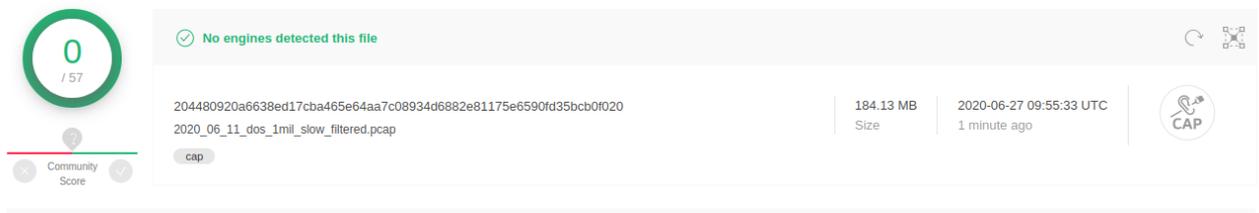

*Figure 4 VirusTotal pcap analysis results*

However, it may be possible that the simulated attacks were not detected by VirusTotal, Snort or Zeek, as they were simply not "intense" enough to reach "DoS" requirements. With this assumption in mind, if the restrictions are not set in the *QoS.xml* file, the attacker can use the *publisher* to cause a Denial of Service in the DDS network.

## 4.3 Preprocessing

The CICFlowMeter[13] was used to turn the traffic recordings (pcap files) into a session dataset (csv file), which is represented as a table of 82 columns wide (See Figure 5).

| Flow Dura | Tot Fwd P | Tot Bwd P | TotLen Fw | TotLen Bw | Fwd Pkt L | Fwd Pkt L | Fwd Pkt L | Fwd Pkt L | Bwd Pkt L | Bwd Pkt L | Bwd Pkt L | Bwd Pkt L | Flow Byts | Flow Pkts | Flow IAT N | Flow IAT S | Flow IAT N | Flow IAT N | Fwd IAT N | Fwd IAT N |
|---|---|---|---|---|---|---|---|---|---|---|---|---|---|---|---|---|---|---|---|---|
| 1.15E+08 | 760 | 0 | 77780 | 0 | 136 | 16 | 102.3421 | 22.07208 | 0 | 0 | 0 | 0 | 673.5704 | 6.581557 | 152139.9 | 590844 | 5109309 | 11 | 1.15E+08 | 152139.9 |
| 1.15E+08 | 313 | 0 | 19984 | 0 | 64 | 16 | 63.84665 | 2.71312 | 0 | 0 | 0 | 0 | 173.0604 | 2.710564 | 370109.3 | 901419 | 5185254 | 151 | 1.15E+08 | 370109.3 |
| 1.15E+08 | 758 | 0 | 77644 | 0 | 136 | 16 | 102.4327 | 22.03042 | 0 | 0 | 0 | 0 | 673.5279 | 6.57532 | 152284.8 | 591413.5 | 5110772 | 13 | 1.15E+08 | 152284.8 |
| 1.15E+08 | 311 | 0 | 19856 | 0 | 64 | 16 | 63.84566 | 2.72183 | 0 | 0 | 0 | 0 | 172.2422 | 2.697791 | 371869.4 | 907136.1 | 5201565 | 112 | 1.15E+08 | 371869.4 |
| 1.15E+08 | 756 | 0 | 77508 | 0 | 136 | 16 | 102.5238 | 21.98809 | 0 | 0 | 0 | 0 | 671.8493 | 6.553105 | 152801.5 | 593431.9 | 5162291 | 5 | 1.15E+08 | 152801.5 |
| 1.15E+08 | 309 | 0 | 19728 | 0 | 64 | 16 | 63.84466 | 2.730625 | 0 | 0 | 0 | 0 | 171.0041 | 2.67844 | 374563.9 | 917090.9 | 5270491 | 130 | 1.15E+08 | 374563.9 |
| 1.15E+08 | 754 | 0 | 77340 | 0 | 136 | 16 | 102.5729 | 21.98105 | 0 | 0 | 0 | 0 | 670.4231 | 6.536062 | 153200.5 | 592846.8 | 5131281 | 17 | 1.15E+08 | 153200.5 |
| 1.15E+08 | 308 | 0 | 19664 | 0 | 64 | 16 | 63.84416 | 2.735054 | 0 | 0 | 0 | 0 | 170.4568 | 2.669888 | 375767.6 | 915530.5 | 5302739 | 107 | 1.15E+08 | 375767.6 |
| 1.15E+08 | 754 | 0 | 77340 | 0 | 136 | 16 | 102.5729 | 21.98105 | 0 | 0 | 0 | 0 | 670.1917 | 6.533806 | 153253.4 | 593800.1 | 5118082 | 22 | 1.15E+08 | 153253.4 |
| 1.15E+08 | 308 | 0 | 19664 | 0 | 64 | 16 | 63.84416 | 2.735054 | 0 | 0 | 0 | 0 | 170.3954 | 2.668927 | 375902.8 | 914599.1 | 5251312 | 104 | 1.15E+08 | 375902.8 |
| 1.15E+08 | 757 | 0 | 77544 | 0 | 136 | 16 | 102.4359 | 22.04481 | 0 | 0 | 0 | 0 | 672.0885 | 6.561062 | 152616 | 589123.7 | 5109331 | 16 | 1.15E+08 | 152616 |
| 1.15E+08 | 311 | 0 | 19856 | 0 | 64 | 16 | 63.84566 | 2.72183 | 0 | 0 | 0 | 0 | 172.0944 | 2.695475 | 372188.8 | 904645.5 | 5236055 | 139 | 1.15E+08 | 372188.8 |
| 1.15E+08 | 753 | 0 | 77272 | 0 | 136 | 16 | 102.6189 | 21.95945 | 0 | 0 | 0 | 0 | 671.7216 | 6.545791 | 152973.1 | 592740.8 | 5127880 | 26 | 1.15E+08 | 152973.1 |
| 1.15E+08 | 307 | 0 | 19600 | 0 | 64 | 16 | 63.84365 | 2.739505 | 0 | 0 | 0 | 0 | 170.373 | 2.668597 | 375953.4 | 915426.1 | 5241811 | 137 | 1.15E+08 | 375953.4 |

*Figure 5 Snapshot of an output of CICFlowMeter*

Since the attack originates from a single IP address, the process of labelling malicious and benign sessions can be automated. As the attack originates only from the 10.0.5.6 VM, a script was composed to mark the networks sessions that contain "10.0.5.6" in IPsrc, IPdrs, or both.

---

[11] Snort, last accessed 2020-06-15, https://www.snort.org/
[12] Zeek (previously known as Bro), last accessed 2020-06-15, https://zeek.org/
[13] University of Brunswick, last accessed 2020-05-01, https://www.unb.ca/cic/research/applications.html ; https://github.com/CanadianInstituteForCybersecurity/CICFlowMeter

Once the full dataset is composed, some of the features have to be cleaned, unified, and engineered, while sessions have to be labelled.

### 4.3.1 Feature extraction and engineering

For this experiment, all feature selection functions were applied to the same dataset with the same shuffle, to ensure the identical environment for performance evaluation.

The network traffic contained several features, that cannot be processed by the neural network, as they are not *float* values, such as timestamps, IP addresses, session IDs, etc. After the *.pcap* file is converted into a *csv* file, a few alterations had to be made for all the values to be encoded as those numeric *float* values.

The "session IDs" were removed from the dataset completely, as being non-representative for the ML process. Also, for the clarity of the experiment, the dataset had the non-DDS sessions removed. Thus, any traffic in the dataset was only the DDS traffic.

Values like timestamps and IP addresses were substituted with the numeric values, as in their initial state DNN was not able to analyse them.

#### *4.3.1.1 Timestamps*

After having been processed with the CICFlowMeter, the timestamps were presented in the format *"day/month/year hour:minute:second AM¦PM"*. For the purpose of the experiment, timestamps were transformed into seconds, and then recalculated relative to the earliest session, which was marked as "0". The rest of the sessions had timestaps as delta between the time of the session and the session right before it.

#### *4.3.1.2 IP addresses*

All the devices were implied to be in the same subnetwork, thus, instead of converting the IP addresses into a decimal format, they were converted into integer numbers by preserving the only address point that varies (e.g., 10.0.5.5 -> 5).

As the IP addresses indicate the origin and the destination of traffic, for some experiments the source and destination IPs are changed in various ways or completely removed, in order to test the importance of addresses as features.

##### 4.3.1.2.1 Randomisation of the IP addresses

The IP addresses were randomly selected from a range in a subnetwork, shuffled and assigned to the sessions randomly.

Randomisation of the IP addresses created confusion in the Clone and MalSub attack detection, resulting in a very slow learning rate. The final accuracy was at 50-55%.

| **Before** | 10.0.5.2 | 10.0.5.3 | 10.0.5.4 | 10.0.5.5 | <u>10.0.5.6</u> |
|---|---|---|---|---|---|
| **After** | 10.0.5.4 | <u>10.0.5.6</u> | 10.0.5.5 | 10.0.5.2 | 10.0.5.3 |

##### 4.3.1.2.2 Shifting IP addresses

While preserving a sequence of virtual machines and routers, the IP addresses from the dataset were shifted by one or more steps. For example, first in a small subnetwork, 10.0.5.2 became 10.0.5.3, while the last 10.0.5.6 became 10.0.5.2.

| **Before** | 10.0.5.2 | 10.0.5.3 | 10.0.5.4 | 10.0.5.5 | <u>10.0.5.6</u> |
|---|---|---|---|---|---|
| **After** | <u>10.0.5.6</u> | 10.0.5.2 | 10.0.5.3 | 10.0.5.4 | 10.0.5.5 |

As long as genuine *publishers* (10.0.5.5) and malicious *publishers* (10.0.5.6) do not overlap, the attacks were detected without any change, with a high accuracy.

#### 4.3.1.2.3 Switching IP addresses

The recorded IP addresses switched places. For example, the attack machine had IP 10.0.5.6 and the genuine publisher had IP 10.0.5.5. They switched places in the test dataset.

The model failed to train and detected malicious traffic as genuine, while the genuine traffic was identified as malicious.

| **Before** | 10.0.5.2 | 10.0.5.3 | 10.0.5.4 | 10.0.5.5 | <u>10.0.5.6</u> |
| **After**  | 10.0.5.2 | 10.0.5.3 | 10.0.5.4 | <u>10.0.5.6</u> | 10.0.5.5 |

With the IP addresses present in the dataset, the model relies on them as main features (see more in section 4.6).

### 4.3.2 Dataset labelling

Labelling of the dataset was performed in multiple ways to avoid marking discovery traffic as malicious, since the discovery traffic can be both benign and malicious. Marking traffic for both directions leads to an increased number of false positives and false negatives during the detection process.

With this assumption, the directional labelling was used for those sessions with the IP address as *10.0.5.6, namely*:
- Bidirectional: only for DoS;
- Only destination IP address: for DoS and Clone;
- Only source IP address: for DoS.

The IP "10.0.5.6", assigned to VM3, was marked as malicious ("DoS", "Clone", or "MalSub") using an automatic script that detects if the IP is present in the session as Source or Destination. Then the "benign" label was replaced with an appropriate attack label.

As in the case of "Malicious Subscriber", the traffic was recorded from the side of the *publisher*, the dataset contained additional IP addresses of the *virtual routers* 10.0.5.2 and 10.0.5.3, which were not present during the simulations of DoS and Clone attacks. Thus, the sessions with those IP addresses had to be removed from the dataset.

### 4.3.3 Automatic feature selection

With all three attacks different in behaviour, the feature selection had to be performed based on the attack type.

To analyse the impact of features on the output of the neural network, the IP addresses were removed for privacy preservation, as well as to remove potential biases.

In the detection of the Clone and Malicious Subscriber attacks, the bias was introduced as the "network discovery" traffic was marked as Benign. Meanwhile, sessions with the data from the topics were predicted by the neural network at 0.45 to 0.55, meaning that the neural network was unable to identify sufficient difference between malicious and benign topics.

Following the approach, outlined in section 3.2 Deep Learning and Feature selection, Table 1 shows the outputs of the functions that select the most impactful features in the dataset.

*Table 1 Feature selection using mathematical methods*

| LassoCV | RFE | Univariate | Importance |
|---|---|---|---|
| TotLen Fwd Pkts | Tot Bwd Pkts | Bwd Pkts/b Avg | Idle Mean |
| Fwd Header Len | TotLen Bwd Pkts | Idle Max | Fwd Header Len |
| Pkt Len Var | Fwd Pkt Len Min | Idle Mean | Idle Max |
| Pkt Len Std | Fwd Pkt Len Mean | Idle Min | Timestamp |
| Flow IAT Mean | Bwd Pkt Len Max | Fwd Act Data Pkts | Flow IAT Mean |
| Bwd Pkts/b Avg | Bwd Pkt Len Mean | Tot Fwd Pkts | Fwd Pkts/s |
| Fwd Pkt Len Std | Flow IAT Mean | Fwd Header Len | Fwd IAT Min |
| Subflow Fwd Byts | Flow IAT Max | Subflow Fwd Pkts | Flow IAT Max |
| Flow IAT Std | Flow IAT Min | Tot Bwd Pkts | Fwd Act Data Pkts |
| Flow Byts/s | Fwd IAT Mean | Bwd Header Len | Flow IAT Min |
| Flow IAT Max | Fwd IAT Max | Down/Up Ratio | Fwd IAT Mean |
| Fwd Seg Size Avg | Fwd IAT Min | Subflow Bwd Byts | Flow Duration |
| Fwd Pkt Len Min | Fwd Header Len | TotLen Bwd Pkts | Fwd IAT Max |
| Flow Pkts/s | Bwd Header Len | Bwd Pkt Len Max | Subflow Fwd Byts |
| Fwd IAT Std | Pkt Len Min | Bwd Pkt Len Min | Tot Fwd Pkts |
| Fwd Pkt Len Mean | Down/Up Ratio | Bwd Pkt Len Mean | Bwd Blk Rate Avg |
| Pkt Size Avg | Fwd Seg Size Avg | Bwd Seg Size Avg | Bwd Pkts/b Avg |
| Flow Duration | Bwd Seg Size Avg | TotLen Fwd Pkts | TotLen Fwd Pkts |
| Bwd Header Len | Subflow Fwd Pkts | Bwd Pkts/s | Flow IAT Std |
| Idle Std | Fwd Act Data Pkts | Subflow Fwd Byts | Fwd IAT Tot |

Out of 78 features generated by *CICFlowMeter*, the 20 most impactful features were selected using the automatic methods described in section 3.2. This reduced the dataset by almost four times. For the experimental purposes, the number of features was reduced further to 10 and 5 to seek further improvement in training time of the machine learning models. The detection rate with the limited features is presented in Table 6 in section 4.6.4.

As indicated in Table 1, for the dataset containing all three attack types at once, the features related to the forward-directed packets showed themselves more impactful than backward-directed.

### 4.4 Selection and application of deep learning for attack detection

There are multiple types of DNNs, that have already been applied to the network intrusion detection. Below are the most common applications of the most widely used types of the deep learning models:
- Dense DNN is used for pattern recognition;
- CNN is used for patterns with many elements, such as images or video data;
- LSTM is used for the time-distributed pattern recognition.

CNN was selected for the experiment as it performs better when analysing multidimensional arrays of features using convolution. The shape of the CNN was selected to have 78 input neurons, multiple hidden layers, and one output neuron providing an output value between 0 and 1 in order to detect if the session is *malicious* or *benign*.

With a "trial and error" approach, the experiment concluded that the CNN with less than three hidden layers can analyse the dataset with a very low accuracy rate and does not result in sufficient improvement. On the contrary, with the number of hidden layers exceeding five, the accuracy failed to improve beyond 50%. Thus, the final shape of the CNN was selected to be 78 input neurons, 3-4 hidden layers, and 1 output neuron (see Figure 6).

As the number of output neurons was less that the number of input neurons, to maintain the optimal learning rate, the number of neurons in a selected layer had to be equal or less than the number of neurons in the previous layer, and higher or equal than the number of neurons in the layer after it. For example, a neural network with the shape 78>78>64>39>1 or 78>78>39>18>1.

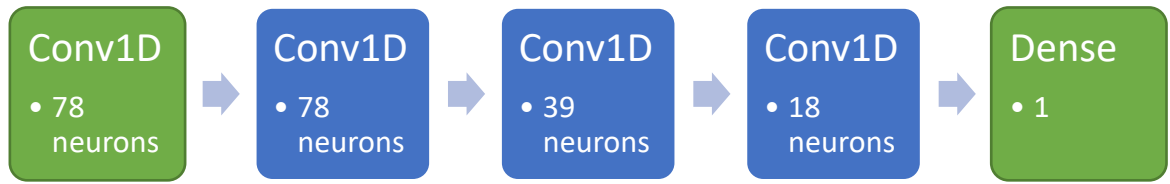

*Figure 6 Shape of the Convolutional Neural Network, used for the experiment*

The shape presented in Figure 6 was further used in all the experiments for the optimal accuracy and detection rate.

### 4.5 Preliminary tests

In the preliminary tests, the dataset contained all the 78 features, except IP addresses, Ports, and Timestamps. This was done to observe the behaviours of the CNNs, and initial accuracy before any modification of the dataset.

The DoS attacks were detected by the first version of the IDS with the 98% accuracy. The missing 2% were False Positive alerts, that were caused by some of the "network discovery" sessions marked as both *benign* and *malicious*.

The Clone attacks were misclassified as benign, causing the accuracy to vary between 50% and 70%. The neural network showed almost all the attack sessions as false negatives, with no True Negatives.

The "Malicious Subscriber" attack detection yielded the 100% false negatives, by recognising the malicious traffic as benign.

For the DoS detection, the maximum accuracy was observed when all the sessions with the Source IP "10.0.5.6" were marked as malicious. The same was true for the Clone attack.

The following observations were also made during the initial tests:
- The source and destination port numbers reduced the accuracy by 2-3% for all three attacks, instead of improving it. Thus, the port numbers had to be removed from the dataset.
- Adding timestamps as a feature improved the results of the "Malicious Subscriber" attack.
- With the source and destination IP addresses present, the detection rate of all the attacks reached 100% and the number of False Positives and False Negatives was almost always 0, provided the attack originates from a new IP address and no genuine (non-malicious) traffic originates from that same IP.

The preliminary experiments allowed to take into account the specificities of the dataset and the DDS environment.

### 4.6 Single neural network vs Ensemble

To further inspect the capabilities of Deep Learning, the performance of a single neural network trained using a full dataset was compared with the simultaneously implemented three types of attacks against an ensemble of three expert CNNs, each trained to detect a dedicated type of attack.

For the sake of experimental comparison, the ensemble adjudication was configured to high sensitivity. In the decision-making process, if at least one expert CNN voted that the session is an attack, the adjudication result would be considered to be an attack. For the session to be considered "benign", all three expert CNNs had to vote "benign".

### 4.6.1 With IP address

To observe the dependence on the origin of traffic, the first experiment was carried out using the dataset that contained both source and destination IP addresses.

*Table 2 Detection results using the dataset with IP addresses present*

|  | TP | FP | TN | FN | Accuracy | Detection rate |
|---|---|---|---|---|---|---|
| DoS | 3482 | 0 | 374 | 0 | 100% | 100% |
| Clone | 3482 | 0 | 780 | 0 | 100% | 100% |
| Malicious Subscriber | 3482 | 0 | 644 | 0 | 100% | 100% |
| **SINGLE CNN** | 3480 | 2 | 1797 | 1 | 99.96% | 99.90% |
| **ENSEMBLE** | 3482 | 0 | 1798 | 0 | 100% | 100% |

With both IP addresses present in the dataset (see Table 2), a single CNN and an ensemble of expert learners achieved very high accuracy within the range of 99.9-100%, relying on the origin of the attack and the key feature. However, a single CNN under the same conditions required twice as much time to train to 99.96% accuracy, as compared to the ensemble of expert CNNs trained to 100% accuracy.

To explore the detection rate against the dataset with limited knowledge of traffic origin, the two experiments were conducted by removing either source IP address or destination IP address.

### 4.6.2 IP Source

By removing the destination IP, the training took more time and the maximum accuracy was limited and never gone higher than 84% for a single neural network detecting multiple types of attacks (see Table 3).

*Table 3 Detection results using the dataset with destination IP addresses removed*

|  | TP | FP | TN | FN | Accuracy | Detection rate |
|---|---|---|---|---|---|---|
| DoS | 3479 | 3 | 364 | 10 | 99.66% | 97.32% |
| Clone | 3278 | 204 | 598 | 182 | 90.94% | 76% |
| Malicious Subscriber | 3480 | 2 | 334 | 310 | 92.43% | 51.8% |
| **SINGLE CNN** | 3386 | 24 | 1004 | 866 | 83.10% | 53.68% |
| **ENSEMBLE** | 3482 | 0 | 1490 | 308 | 94.17% | 82.86% |

The detection rate of both methods was significantly lower than in the previous experiment with the source and destination IPs available. The detection rate in the single CNN dropped down to almost 50%. Similar to the case in Section 4.6.1, the CNNs heavily rely on the IP address as a feature.

### 4.6.3 IP Destination

After removing the destination IP addresses from the full dataset, the results were similar to the previous experiment with the source IP addresses removed (Table 4).

*Table 4 Detection results using the dataset with source IP addresses removed*

|  | TP | FP | TN | FN | Accuracy | Detection rate |
|---|---|---|---|---|---|---|
| DoS | 3482 | 0 | 355 | 19 | 99.51% | 94.91% |
| Clone | 3479 | 3 | 683 | 97 | 97.65% | 87.56% |
| Malicious Subscriber | 3472 | 10 | 332 | 312 | 91.19% | 51.00% |
| **SINGLE CNN** | 3042 | 454 | 1654 | 130 | 88.9% | 92.71% |
| **ENSEMBLE** | 3430 | 52 | 1462 | 335 | 92.67% | 81.36% |

As before in the experiment in section 4.6.1, the CNNs relied heavily on the IP address as main features. The detection rate was lower than in the full dataset. However, it was higher than in the experiment in section 4.6.2. The detection rate for a single CNN was over 92%, as compared to almost 50% using the dataset with only destination IP addresses.

### 4.6.4 No IP addresses

To test the final privacy-preserving network intrusion detection system, source and destination had to be removed from the dataset, that contains all three types of attack traffic.

*Table 5 Detection results using the dataset with source and destination IP addresses removed*

|  | TP | FP | TN | FN | Accuracy | Detection rate |
|---|---|---|---|---|---|---|
| DoS | 3482 | 0 | 337 | 37 | 99.04% | 90.10% |
| Clone | 3170 | 312 | 612 | 168 | 88.74% | 78.46% |
| Malicious Subscriber | 3470 | 12 | 30 | 630 | 84.50% | 4.60% |
| **SINGLE CNN** | 3126 | 324 | 672 | 1118 | 71.90% | 37.54% |
| **ENSEMBLE** | 3442 | 40 | 1074 | 724 | 85.53% | 59.73% |

The accuracy without IP addresses significantly reduced (see Table 5). However, this approach allows detection of attacks that originate from any machine without prior knowledge of their origin and destinations. This approach ensures privacy and takes into account that some of the attacks might originate from the same device or from a different subnetwork, while still being detected by the IDS.

The distribution of the CNN predictions (see Figure 7) illustrates that the model was able to classify the majority of the analysed sessions with only 60% certainty. Only the DoS attack sessions and a very small amount of benign traffic were identified with 100% certainty.

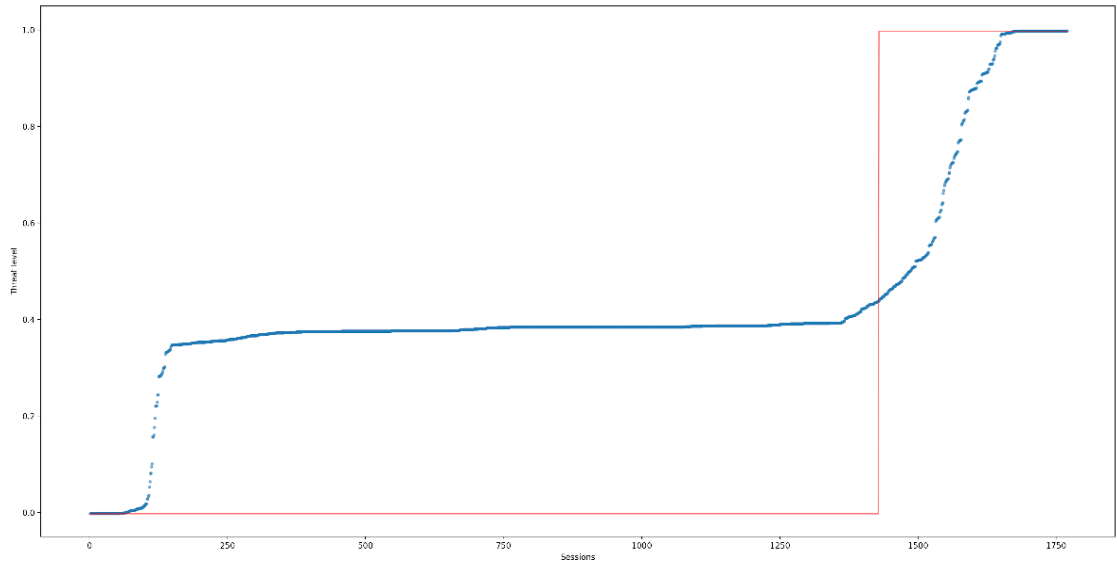

*Figure 7 Detection distribution of a single universal CNN using the dataset without IP addresses (red – expected; blue - predicted)*

With further training the distribution of predictions did not change. The assumption was made that the issue of the majority of samples being classified with a low certainty is due to insufficient difference between benign and malicious samples.

Following the feature selection in the experiment in section 4.3.3, the same training was performed using only the features with high impact (see Table 6).

*Table 6 Detection rate using the dataset without IP addresses with only the feature with high impact*

| Number of Features | DoS | Clone | MalSub | Single CNN | Ensemble |
|---|---|---|---|---|---|
| 5 | 90% | 27% | 0% | 15% | 22% |
| 10 | 97% | 53% | 1% | 20% | 31% |
| 20 | 100% | 80% | 2% | 29% | 43% |

The automatic feature selection allowed to improve training time by an order of magnitude, while noticeably reducing the detection rate for both single universal CNN and the ensemble of expert learners. The detection rate was insufficient for the intrusion detection to be a viable software solution. Based on the findings the following conclusions were made:

1. DoS attack required only the most impactful five features to be detected with the accuracy of 90%.
2. Clone attack required at least 20 features to be detected – key features were the Average packet rate, Average backward packet rate, Average Bulk Forward rate and Protocol. It required up to 20 features to ensure similar detection rate as with all 78 features.
3. "Malicious Subscriber" attack required a full set of 78 features. However, even this may be insufficient to detect the attack with a good accuracy. Thus, the automatic feature selection method is not suitable for this type of attack.

The potential use of such intrusion detection system, its drawbacks and limitations, as well as other defence solutions will be discussed in the following section.

# 5   Discussion

The experiments have shown a high potential of Machine Learning to be used for the network intrusion detection in DDS while preserving privacy of the content. Though with a relatively low detection rate, the CNN was able to detect anomalies in the network sessions without accessing potentially sensitive data and without the reliance on the behavioural analysis of the network traffic.

The findings are based on the three main types of attacks, that can be delivered through the DDS protocols and APIs:
- Denial of Service
- Clone
- "Malicious Subscriber"

Among these three types, the DoS attack is generally the easiest to detect and prevent. Specific to Machine Learning, features like "average packet length", "total length of packets", "packet length mean", allow a relatively high detection rate (up to 90%), without any knowledge of the attack origin or destination.

The other two attack types are harder to detect and require specific conditions to be identified. The Clone attack is an attack that is much harder, or even impossible, to detect using the traffic flow analysis, specifically if the compromised instance broadcasts fake coordinates that still follow the same format and rules. The features, extracted by *CICFlowMeter*, are insufficient to successfully detect the attacks with reliable accuracy. The issue can be partially resolved with the use of the static IP addresses in DDS networks.

The "Malicious Subscriber" attack is detectable under the condition of the IP addresses being present in the dataset and if the attack originates from the same device. Additionally, the number of the received topics by the *subscriber* should be different from the number of topics per the network session, received by a benign *subscriber*. This event is typical in the industrial network, a device with a benign *subscriber* would have a higher uptime, than the device with a malicious *subscriber*.

With the attempts to anonymise the dataset and preserve privacy, the detection rate of all the attacked reduced. Without the IP addresses in the dataset, the detection rate for the "Malicious Subscriber" attack reached only 5%. Detection of all three attacks should involve additional measures, including the rule-based authentication of subscribers.

## 5.1   Highest detection rate and performance

The findings revealed that out of all the experiments meeting the criteria of privacy preservation, the best detection rate was achieved by an ensemble of the expert CNNs, where each network was trained to detect only one type of attack using the full dataset with 78 features. With a basic adjudication of predictions, if at least one of the expert networks suspected an attack, the attack was acknowledged. The accuracy reached 85.53% and the detection rate - 59.73%.

The best training speed was detected for a single universal CNN using the datasets with 5 and 10 automatically selected most impactful features. The difference in training time between "5" and "10" was negligible, thus, the higher detection rate was selected for the analysis. The CNN, trained with the 10-feature dataset reached the detection rate of 20%, which is three times lower than the accuracy of the top performing ensemble of expert learners.

## 5.2   Selected solutions

Considering the variance in the detection rates in the applied methods, it was observed, that for the maximum performance some special conditions have to be met. They may be also considered as the limitations or constraints. For example, using IP addresses and consequently undermining the privacy preservation requirement.

It is highly probable that the performance can be improved through the generation of additional samples. With more features and a larger dataset, CNN can be more sensitive to minor changes in the network behaviour and improve the overall accuracy of the IDS.

Based on the experiment findings, the following security solutions could improve the security of the systems that use DDS:
- Static Rules and Permissions
- Domain grouping
- Sandboxing
- Cleansing

The listed solutions are not exhaustive, but they should provide an adequate level of security against the known attacks.

### 5.2.1 Static Rules and Permissions

Static rules and permissions are configured once and remain unchanged during the exploitation of the system. For DDS, the term "rule-based security" implies restrictive configuration of the *QoS.xml* and other rule files.

RTI Connext DDS itself (and other DDS implementations) has a built-in access to the monitoring data and the connected devices. This functionality allows the malicious actors to avoid the steps of the network enumeration, and to acquire the list of connected devices direction from the DDS application.

Very restrictive permissions have to be set in place at the level of the system software, dependencies and applications. DDS has to operate under a separate user account from the rest of the software, and only specific required actions have to be allowed to that account.

Malicious actors are able to launch unauthorised instances of publishers and subscribers, as there is no authentication in the peer-to-peer API. The use of instance IDs in combination with limiting the number of connections, originating from a single device (e.g., health monitors, drone, etc.), can improve the rule-based security.

### 5.2.2 Domain grouping

The subscribers and publishers can be grouped in domains using *QoS.xml* file, in order to be able to publish specific topics for the designated groups of subscribers.

When naming domains, similar to any practice, Domain IDs should be hard to guess, to avoid contamination of other domains in case a malicious actor gains access to a publisher. Hashing IDs might provide a reasonable security, while using an industrial naming scheme for easier maintenance.

### 5.2.3 Sandboxing

Sandboxing of web interfaces and publisher/subscriber executables can be used to enhance security through the isolation of DDS elements from the operating system and other software.

This way only DDS itself is compromised, potentially allowing the malicious actors to propagate inside the DDS network, but not with any other services.

However, sandbox escaping is a known type of attack, which has to be mitigated by other approaches.

Furthermore, the use of virtual containers leads to an additional layer of sophistication and to an increased resource consumption. It can be critical in the industrial setting, as additional computational resources might increase heat, produced by the devices and general power consumption of the facility.

### 5.2.4 Cleansing

The cleansing method is an approach to completely overwrite the software and the operating system, with a pre-configured non-infected image or container, thus, removing any potential malware that might have infected the device without the need to detect it.

Implementation-wise, at the software and operating system levels, virtual containers can be used to perform a faster cleansing, to avoid reinstallation of operating systems and configurations. Full system rewrites can be used as an additional stage of cleansing on less regular bases, even at random intervals.

Cleansing takes time and additional resources, thus, in time-sensitive and mission-critical environment, it could be viable to perform cleansing during maintenance or recharge.

Other security solutions can be explored through further experiments, that may also reveal unknown variables in the process of privacy-preserving intrusion detection in DDS using Deep Learning.

## 6 Conclusion

This report covers the critical issue of vulnerabilities in the DDS implementations (e.g., RTI Connext DDS, OpenSplice, etc.) and their protection with the use of Machine Learning (specifically Deep Learning) methods.

A series of experiments was conducted with the aim to explore the conditions of the detection of the attacks over DDS protocols using a privacy-aware convolutional neural network.

The detection rate proved to be higher than using conventional opensource intrusion detection systems. It is suboptimal for a machine-learning-based IDS, but it can be improved with further research.

The conducted experiment verified three types of the DDS vulnerabilities, and provided insight into their detection feasibility and mitigation measures. With the privacy-preservation dataset, the maximum accuracy reached by an ensemble of expert deep neural networks is 85.53% and the detection rate - 59.73%.

The experiment confirms that cyber attacks remain a strong threat to DDS. Among the selected three types of attacks, the "Malicious Publisher" proved to be the hardest to detect with the current preprocessing and Machine Learning approaches, with a possibility for the DoS attack to be recognized through conventional methods. The "Malicious Subscriber" attacks showed themselves as unique and are almost indistinguishable from the benign traffic.

Deep learning shows high potential in detection of simulated attacks via DDS API using only metadata, with the limited or even no knowledge of the traffic origin. However, to further improve and potentially maximise the security of systems and networks, it is recommended to use a combination of conventional signature-based detection and Machine Learning anomaly detection.

Due to the promising potential of the technology and its wider applicability, DDS requires further research and enhanced security. In the future research, additional types of attacks may reveal additional approaches to securing the DDS at runtime beyond static rules and conventional IDS.

## 7 Acknowledgement

The author would like to acknowledge that the research was conducted under the direct supervision of Dr Peter Popov and funded by the project *Aggregated Quality Assurance for Systems* (AQUAS).